%% file: main.tex
\documentclass[runningheads, anonymous]{llncs}
\usepackage[T1]{fontenc}
\usepackage{amsmath}
\usepackage{amssymb}
\usepackage{graphicx}
\usepackage{xcolor}
\usepackage{rotating}
\usepackage{url}
\begin{document}
\title{QwenSafe: Multimodal Content Rating Description Identification via Preference-Aligned VLMs}
\titlerunning{QwenSafe}
%
\author{Dishanika Denipitiyage\inst{1}\orcidID{0000-0002-8717-098X} \and
Aruna Seneviratne\inst{2}\orcidID{0000-0001-6894-7987} \and
Suranga Seneviratne\inst{1}\orcidID{0000-0002-5485-5595}}
\authorrunning{D. Denipitiyage et al.}
%
\institute{University of Sydney, Sydney, Australia \\
\email{\{dishanika.denipitiyage,suranga.seneviratne\}@sydney.edu.au} \and
University of New South Wales, Australia \\
\email{a.seneviratne@unsw.edu.au}}
\maketitle              
\begin{abstract}
Mobile app marketplaces require developers to disclose standardized content rating descriptors (CRDs) to inform users about potentially sensitive or restricted content. Ensuring the accuracy and consistency of these disclosures remains challenging due to the multimodal nature of app content, which spans textual descriptions and visual interfaces. In this paper, we present QwenSafe, a Vision–Language Model (VLM) designed to automatically identify the presence of Apple-defined CRDs by jointly reasoning over app metadata and screenshots. To enable scalable training for this task, we introduce metadata2CRD, a data-construction pipeline that synthesizes descriptor-aligned question–answer pairs by combining app descriptions, screenshots, and formal descriptor definitions. We adapt Qwen3-VL-8B using supervised fine-tuning followed by Direct Preference Optimization (DPO) to align model predictions with descriptor-specific evidence and explanations across visual and textual modalities. We evaluate QwenSafe on 12 Apple-defined content rating descriptors and compare it against state-of-the-art vision– language models, including Qwen3-VL, LLaVA-1.6, and Gemini-2.5-Flash. QwenSafe consistently outperforms all baselines in binary CRD classification, achieving improvements in positive-class recall of 111.8\%, 36.1\%, and 2.1\%, respectively. Our results demonstrate that descriptor-aware multimodal alignment substantially improves automated content classification and highlights the potential of vision–language models to support scalable and consistent content rating in mobile app marketplaces.\\ \vspace{-3mm}



\end{abstract}

\input sections/1_Introduction
\input sections/2_Related
\input sections/3_Datasets

\input sections/4_Methodology

\input sections/5_Results

\input sections/6_Conclusion
\bibliographystyle{splncs04}
\bibliography{biblio}
\appendix
\input sections/7_appendix
\end{document}

%% file: sections/1_Introduction.tex
\section{Introduction}
\label{chap3_intro}


Mobile applications have become deeply embedded in everyday life, serving both adult and child users at an unprecedented scale. With an average of 1,863 new applications added daily to Apple’s App Store~\cite{appleAppsperday} and 1,617 to Google Play~\cite{androidAppsperday}, users are continuously exposed to a rapidly evolving ecosystem of digital services. In this setting, inaccurate or inconsistent content introduces a security vulnerability, as mobile platforms rely heavily on developer-provided metadata to enforce safeguards such as age-based access control and content restrictions~\cite{chen2013isthisapp}. This reliance creates an opportunity for adversarial developers to intentionally misrepresent application content, enabling policy evasion and bypassing platform protections~\cite{denipitiyage2025detecting,canada_cr_article,sun2023not}. In profit-driven environments, such strategic misclassification can allow applications to reach broader or unintended audiences, exposing users—particularly children—to harmful or inappropriate content. Consequently, these practices undermine the integrity and trustworthiness of mobile ecosystems, highlighting the need for robust, security-oriented mechanisms for content classification and enforcement.

The two leading app marketplaces, Google Play Store and Apple App Store, have implemented multiple safeguards to protect users, including strict developer policies, inspection and vetting procedures prior to publication, providing essential app information (e.g., download counts, permissions, ratings, developer details, and community reviews), and the assignment of age-appropriateness ratings to guide user decision-making. Android leverages the International Age Rating Coalition (IARC)~\cite{iarc} framework, wherein a standardised questionnaire is used to automatically generate region-specific official age ratings for applications, which are subsequently displayed on the Google Play Store. 
In contrast, Apple employs a centralised rating framework across all regions, along with regional exceptions. These rating schemes assess the presence of sensitive content descriptors such as mature themes, sexuality or nudity, violence, gambling, drugs and language~\cite{applecr_descriptors,androidcr_descriptors}. The initial evaluation is based on questionnaires completed by app developers when submitting their apps, where Apple employs a proprietary questionnaire, while Android utilises the IARC framework.
In 2017, Google employed text analysis, image understanding, and static and dynamic analysis of the APK binary to identify potential threats an app poses to its users~\cite{android_mlresearch}. More recently, in 2023, Google introduced machine learning based initial screening to scan apps for policy violations, which reduces the manual workload and filters bad actors before they are ever seen by human reviewers~\cite{android_ml_appreview,android_ml_appreview2}. However, the reliance on self-reported questionnaires and heterogeneous platform-specific review pipelines creates opportunities for adversarial manipulation, where developers can intentionally misclassify or under-report sensitive content to bypass age restrictions and reach broader user populations.

Despite Google's efforts to prevent the publication of apps that violate content rating schemes, a significant 19.25\% of content rating inconsistencies were still observed across different agencies~\cite{sun2023not}. Additionally, 81.25\% of apps categorised as “family-friendly”~\cite{android_family} were found to embed trackers, despite such practices being prohibited for children's apps~\cite{sun2023not}. The Canadian Centre for Child Protection (C3P) found that Google Play often assigns lower age ratings (e.g., “Teen”) than the Apple App Store (e.g., “17+”), allowing younger users access to apps such as YouTube, KIK, Whisper™, and Yubo. They also note that Google Play provides less detailed content descriptors, limiting its descriptors to interactive elements and in-app purchases, while Apple includes more comprehensive content descriptors~\cite{canada_cr_article}.
Furthermore, the C3P states that Apple uses content descriptors similar to the ESRB~\cite{esrb}. Prior work has demonstrated significant inconsistencies in age-rating decisions across platforms, highlighting a broader measurement gap in how mobile app content is assessed and operationalised at scale.

Therefore, there is a pressing need for a unified and security-oriented measurement framework that can systematically assess content risk across mobile applications and mitigate inconsistencies arising from heterogeneous, self-reported rating systems. Such a system would reduce opportunities for adversarial misrepresentation of app content, improve transparency in cross-platform content moderation, and strengthen the reliability of age-based access control mechanisms.
In this paper, we introduce an automated multimodal security measurement framework for fine-grained content rating descriptor (CRD) inference from app metadata. We construct a unified descriptor taxonomy and generate descriptor-aligned training data integrating textual descriptions, screenshots, and formal descriptor definitions. We adapt the Qwen3-VL-8B model, termed \textit{QwenSafe}, as a multimodal security classifier for CRD reasoning. QwenSafe consists of two components: (1) metadata2CRD, which transforms multimodal inputs into structured CRD question–answer pairs, and (2) an offline alignment stage using supervised fine-tuning and Direct Preference Optimisation (DPO)~\cite{rafailov2023direct} to improve robustness in descriptor-specific reasoning under noisy and potentially adversarial inputs.
Our contributions are as follows:
\begin{itemize}
\item We develop \textbf{QwenSafe}, a multimodal security-oriented model adapted for CRD identification, supported by \textbf{metadata2CRD}, a data construction pipeline that produces descriptor-aligned Q\&A pairs by combining app descriptions, screenshots, and formal descriptor definitions.

\item We apply \textbf{supervised fine-tuning} and \textbf{DPO-based preference alignment} to Qwen3-VL-8B, enabling the model to identify descriptor presence and explain relevant cues across visual and textual modalities.

\item Through evaluation on 12 Apple-defined descriptors, QwenSafe outperforms Qwen3-VL, LLaVA-1.6, and Gemini-2.5-Flash, improving positive class recall by \textbf{111.8\%}, \textbf{36.1\%}, and \textbf{2.1\%}, respectively, in binary CRD classification.  

\end{itemize}

%% file: sections/2_Related.tex
\section{Background}


\subsection{Legal Background and App Market Policies}
\label{chap3_legal_background}
\textbf{Legislation:} This section describes the existing legal/regulatory implementations such as the European General Data Protection Regulation (GDPR), the Children’s Online Privacy Protection Act (COPPA), App Store and Google Play Policies around mobile app content rating, age assurance, and protecting minors from harmful content. Further, we specifically outline recent changes in mobile app content rating in Australia, as this research was conducted by setting the geographical location as Australia.  

\begin{figure}[t]
    \centering
    \includegraphics[width=0.9\linewidth]{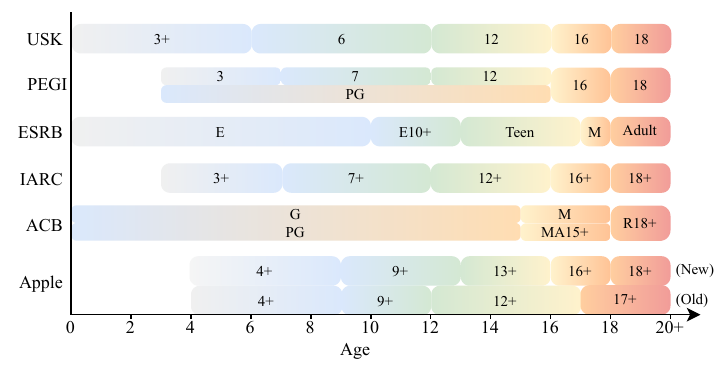}
    \caption{Comparison of age ratings across major authorities (USK, PEGI, ESRB, IARC, ACB, and Apple)}
    \label{fig:cr_authorities}
    \vspace{-0.5cm}
\end{figure}

Children need constant vigilance and effort to protect their personal data, as they often do not fully understand the risks of how their data is collected and used. According to COPPA §312.4~\cite{coppa_312_4}, applications directed at users under the age of 13 are required to make reasonable efforts to ensure that a child’s parent receives direct notice of the app’s practices concerning the collection, use, or disclosure of personal information. Similarly, under GDPR~\cite{gdpr} Art. 8, the processing of a child’s ( below the age of 16 years) personal data is lawful only when consent is provided by the holder of parental responsibility and Art. 5 requires that user personal data be processed “lawfully, fairly and in a transparent manner in relation to the data subject.” These regulatory efforts collectively protect the underage audience from intentional or accidental access to age-restricted applications. While reputed organisations strictly follow such regulatory guidelines and reflect the target audience via content rating labels, there is no guarantee that not-on-spotlight developers are representing the actual content of their apps via content-rating labels~\cite{banned_apps}. Enforcement actions for such apps, therefore, are reactive in nature, as the regulatory efforts are predominantly focused on restricting access and securing personal information handling of children rather than verifying the content's suitability.

In Australia, the Online Safety Act 2021 (Cth), Part 9~\cite{osa_2021,esafety_com} consolidated powers of the eSafety Commissioner to compel removal and restrict access of harmful mobile apps, complementing the Broadcasting Services Act 1992 (Cth)~\cite{broadcasting_act_1992} and its classification regime. The subsequent Online Safety Amendment (Social Media Minimum Age) Act 2024 introduced Part 4A, establishing a statutory minimum age of sixteen for users of designated social media platforms~\cite{social_media_minimum_age}. By requiring providers to take reasonable steps to prevent underage users from creating or maintaining accounts, providers to purge existing accounts belonging to children, the amendment seeks to reduce children’s exposure to harmful or age-inappropriate content. This law is intended to apply to companies such as Facebook rather than companies offering education and health support. 

\noindent\textbf{Age rating:} Age rating, also referred to as content rating or maturity rating, is defined as evaluating the appropriateness of television programs, films, comic books, video games or mobile apps for their intended audience\cite{pegi,esrb,motionpic} against the availability of different content descriptors. In Android apps, the developers and the International Age Rating Coalition (IARC) are responsible for assigning age ratings. Developers complete a questionnaire provided by the IARC, which then automatically generates region-specific ratings for different parts of the world.
Figure~\ref{fig:cr_authorities} shows the major regional content rating authorities, including the Entertainment Software Rating Board (ESRB)~\cite{esrb} in the Americas, the Pan European Game Information (PEGI)~\cite{pegi} in Europe and the Middle East, the Unterhaltungssoftware Selbstkontrolle (USK)~\cite{usk} in Germany, the Australian Classification Board (ACB)~\cite{acb} in Australia, and the International Age Rating Coalition (IARC)~\cite{iarc}. In Australia, game apps follow ACB ratings (General (G), Parental Guidance (PG), Mature (M), Mature Accompanied (MA15+), and Restricted (R18+)) and the generic apps follow IARC ratings (3+, 7+, 12+, 16+, and 18+). 

In contrast, Apple App Store uses its own rating scheme~\cite{applecr_descriptors,cr_apple_setup}. 
With the release of iOS26~\cite{apple_new_cr_update}, Apple further streamlined the setup process of child accounts—mandatory for users under 13 and available for all individuals up to the age of 18. This update enables parents to authorise the sharing of a child’s age range with app developers, without disclosing the child’s exact date of birth. In parallel, Apple revised its age-rating scheme~\cite{applecr_descriptors} by introducing additional categories (13+, 16+, and 18+) alongside the existing 4+ and 9+ tiers, while eliminating the former 12+ and 17+ ratings. To support this transition, a new developer questionnaire in App Store Connect was also deployed, encompassing expanded content descriptors such as medical or wellness information, violent material, and the availability of in-app control. However, in this research, we focus on Apple's previous content rating scheme, as our dataset was crawled before the changes.

\begin{figure}[t]
    \centering
    \includegraphics[width=0.85\linewidth]{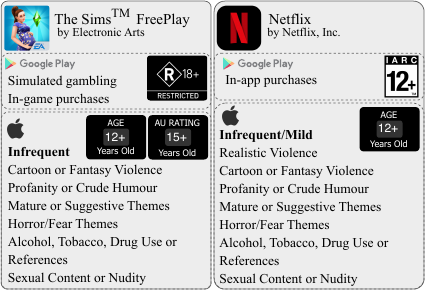}
    \caption{Content descriptors of  $The~Sims^{TM}~FreePlay$ and $Netflix$ apps across App Store and Play Store.}
    \label{fig:sims_freeplay}
\vspace{-0.68cm}
\end{figure}

\noindent\textbf{Content descriptors:} Content descriptors are indicators accompanying age ratings that specify the nature of potentially sensitive or objectionable material present in an app. These indicators are defined by a regulatory body or by the app market itself. For example, ACB guidelines use a hierarchical descriptor scheme where the first layer has eight different indicators, themes, violence, sex, language, drug use, nudity, online interactivity, and in-game purchases, and assess the impact to decide the age rating. More specifically, Android follows five Likert-style scales: ‘very mild’, ‘mild’, ‘available’ and ‘strong’ and ‘very strong’. iOS apps follow a similar structure containing seven main descriptors (In-App Controls, Capabilities, Mature Themes, Medical or Wellness, Sexuality or Nudity, Violence, Horror/Fear Themes,
Chance-Based Activities) and consider ‘mild/infrequent’ or ‘intense/frequent’ presence of the descriptor. Even though both app markets have different first layers in their hierarchy, they both follows almost similar descriptors in the second layer. However, ACB provides fine granular second layer compared to apple. For example ACB divides violence into 12 sub-categories whereas Apple has only four sub-categories. Compared to Android, Apple recently introduces Parental Controls and Age Assurance as a protection layer for children under 16 years. 

\subsection{iOS and Android Content Rating Schemes}

Content ratings and descriptors can diverge substantially even for the same application when distributed across iOS and Android, owing to the fact that developers respond to different platform-specific questionnaires during the submission process. As illustrated in Figure~\ref{fig:sims_freeplay}, \emph{$The ~Sims^{TM}$ FreePlay} by \emph{Electronic Arts} receives an R18+ rating on Android but only 15+ on iOS, despite representing the same title with identical screenshots. The underlying descriptors further highlight this discrepancy: Android attributes its classification to simulated gambling and in-game purchases, whereas iOS cites sexual content and nudity, alcohol and tobacco use, mature or suggestive themes, and several additional factors. Such inconsistencies raise a fundamental question about how two rating schemes can govern an identical game within the same region and which set of descriptors more accurately reflects the app’s content. This phenomenon extends beyond games. For instance, \emph{Netflix} is rated 12+ on both platforms, yet the rationale differs markedly: Android points solely to in-app purchases, while iOS lists realistic violence, profanity or crude humour, mature or suggestive themes, sexual content and nudity, and several others. More concerning is that developers appear to treat each app marketplace as a distinct regulatory environment, even when distributing the same product. In practice, households often operate a mix of Android and iOS devices, leaving end users—particularly parents—uncertain about which rating or set of descriptors to trust when evaluating apps they have no prior familiarity with. 


Additionally, we observed several content rating descriptors that are visible only in certain app ecosystems. In iOS, if developers claim their game contains infrequent nudity or sexual content, that app may no longer receive a content rating less than 12+. In Android, very mild nudity or sexual content could be present in a general (G) rated game, and more occurrences require parental guidance. It is again concerning how Australian regulators are not binding both behaviours to avoid this confusion to both app developers and end-users alike. 
In this work, we aim to reduce this gap by comparing both ecosystems and creating more references among them.


\section{Related Work}
\label{chap3_related}

\noindent \textbf{Content Rating:} 
Comparatively fewer efforts address content rating as a security and measurement problem, where inaccurate or manipulated ratings can enable policy evasion and inappropriate content exposure in mobile ecosystems.
Early work by Chen et al.~\cite{chen2013isthisapp} proposed Automatic Label of Maturity ratings (ALM), a semi-supervised text-mining approach that infers maturity ratings using app descriptions and user reviews, treating Apple App Store ratings as ground truth. However, this method relies primarily on keyword matching and lacks semantic understanding. 
As a result, such approaches are vulnerable to evasion through benign wording or deliberate obfuscation in app metadata, limiting their effectiveness in adversarial settings. Hu et al.~\cite{hu2015protectingcikm} extended this line of work by introducing a text feature–based SVM classifier with online training, but similarly depended exclusively on textual cues. Subsequent studies by Liu et al.~\cite{liu2016identifying} and Chenyu et al.~\cite{zhou2022automatic} incorporated additional modalities, including app icons, screenshots, and APK features, to identify children’s apps. Nevertheless, their feature extraction was limited to surface-level signals such as OCR-extracted text, colour distributions, permissions, and APIs, without deeper cross-modal reasoning. These limitations restrict their ability to detect inconsistencies between declared ratings and actual content.

More recent work by Sun et al.~\cite{sun2023not} analysed inconsistencies in content ratings across geographic regions by mapping rating systems between jurisdictions. Complementing these findings, an investigation by the Canadian Centre for Child Protection Inc. (C3P)~\cite{canada_cr_article} in 2022 reported that app age ratings often vary across Apple’s App Store, Google Play, and the app’s own terms of service, and further noted that both major mobile app stores lack transparency in how age ratings are determined. Beyond automated methods, longitudinal audits have revealed substantial non-compliance with platform and regulatory standards. For example, Xiao~\cite{xiao2025non} reported widespread violations in the UK loot box industry, where apps rated as suitable for children (e.g., 4+ or 9+) nonetheless imposed 18+ age requirements through in-app gating mechanisms. Similarly, Carter et al.~\cite{carter2025investigating} identified systemic failures in Australia’s Mandatory Minimum Classifications Scheme for gambling-like content. 
Collectively, these studies expose a systemic measurement gap between developer-declared ratings and actual app content available, highlighting the need for automated auditing mechanisms capable of identifying misrepresentation at scale.

\noindent \textbf{Multimodal Harmful Content Detection:} Large language models (LLMs) and vision–language models (VLMs) have increasingly been applied to security-relevant content analysis tasks, including the detection of abusive, deceptive, or policy-violating content.
Early work~\cite{chiu2021detecting} evaluated the ability of GPT-3 to identify sexist and racist text, demonstrating that few-shot prompting could achieve accuracies of approximately 85\%. Building on this line of research, Guo et al.~\cite{guo2023investigation} systematically explored four prompting strategies for online hate speech detection using LLMs, reporting substantial F1-score improvements ranging from 7.9\% to 24.2\% on the HateXplain dataset~\cite{mathew2021hatexplain}. Beyond text-only approaches, VLMs extend content understanding by jointly modelling visual and textual signals, enabling the detection of multimodal harmful content. The Hateful Memes Challenge~\cite{kiela2020hateful} introduced a benchmark highlighting the limitations of unimodal models and the necessity of multimodal reasoning. Pro-Cap~\cite{cao2023pro} employed frozen CLIP-based representations to identify hateful meme content, while Vid+RM-FT~\cite{wang2025cross} fine-tuned large-scale VLMs (LLaMA-3.2-11B and LLaVA-NextVideo-7B) on HateMM and re-annotated Hateful Memes datasets \cite{kiela2020hateful}. 

Despite this progress, state-of-the-art LLMs and VLMs often require extensive task-specific fine-tuning, carefully engineered prompting strategies~\cite{guo2023investigation}, or additional supervision to generalise reliably to new domains. 
Moreover, existing work does not address the problem of detecting misrepresentation in structured platform metadata, where developers may intentionally under-report sensitive content.
\emph{To the best of our knowledge, this is the first work to model content rating of mobile apps as a multimodal security auditing problem, leveraging vision–language models to detect inconsistencies between declared and inferred content descriptors in mobile app marketplaces.}
\vspace{-0.3cm}

%% file: sections/3_Datasets.tex
\section{Data Pipeline}
\label{chap3_datasets}


In this section, we formulate our content rating descriptor (CRD) taxonomy creation and dataset curation for CRD prediction task. \\ \vspace{-3mm}

\noindent{\bf Content descriptor taxonomy: }
\begin{figure}[t]
    \centering
    \includegraphics[width=0.95\linewidth]{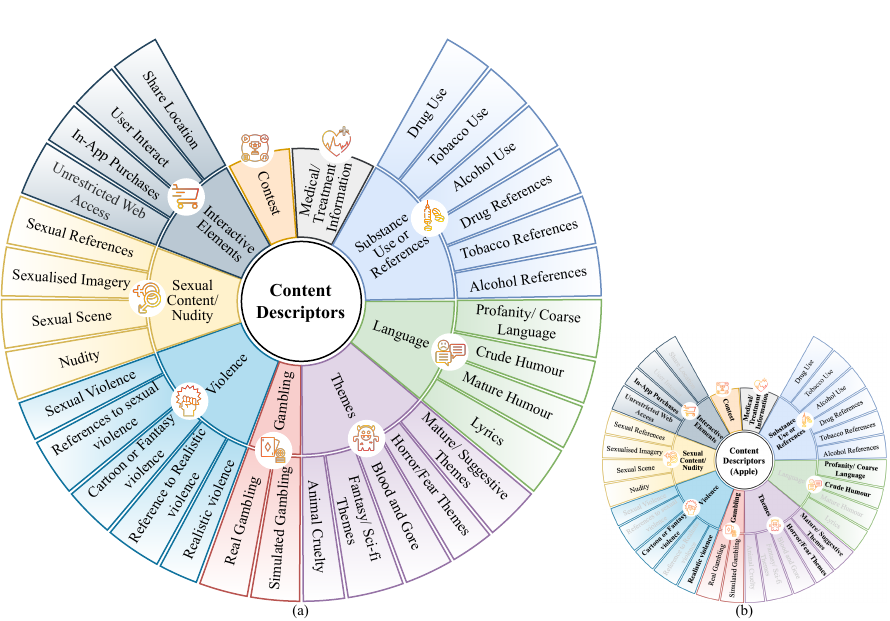}
    \caption{(a) Content descriptor taxonomy and (b) mapping to 12 different Apple content rating descriptors }
    \label{fig:cd_taxonomy}
    \vspace{-0.5cm}
\end{figure}
The CRD taxonomy in Figure~\ref{fig:cd_taxonomy} (a) is constructed by integrating multiple content rating authorities rather than relying on a single framework. Specifically, we analyse descriptor definitions from Apple (iOS) and the Australian Classification Board (ACB), and further compare them with ESRB and PEGI standards to identify both shared and region-specific elements. We define nine primary content categories: Language, Themes, Gambling, Violence, Sexual content and Nudity, Medical Treatment Information, Substance Use, Contest, and Interactive Elements, by extracting concepts common across rating systems while preserving unique dimensions (e.g., “Themes” in ACB capturing elements, such as fear or horror in PEGI and blood or gore in ESRB). Each primary category is further decomposed into 2–6 fine-grained sub-categories, resulting in a total of 30 sub-categories.

This process yields a unified, hierarchical taxonomy that fully covers all 12 content descriptors defined in the iOS ecosystem (\textbf{cf.}~Figure~\ref{fig:cd_taxonomy} (b)) and all first-layer descriptors specified by ACB, while remaining sufficiently abstract to generalise across other rating schemes. As a result, the taxonomy can be readily applied to regional schemes within the Android ecosystem, particularly those based on ESRB and PEGI standards. Its concise yet extensible structure enables flexible representation of content across app marketplaces and supports progressive refinement during data construction and model training. This adaptability is critical for maintaining robustness in the presence of evolving content and potential inconsistencies in mobile app ecosystems.



\noindent{\bf Curation of training and testing datasets: }
The manual reviewing process for iOS apps compared to the automated review in Android has resulted in selecting iOS data as ground truth for verifying Android apps in many prior research~\cite{chen2013isthisapp}. Following this intuition, we construct our training dataset using developer-defined content descriptors, app metadata, and content-rating labels from popular iOS applications. We crawled a large-scale dataset of approximately 1.2 million Apple App Store entries between January and November 2023. From this corpus, we selected the top 780 apps from each content-rating category, excluding the 4+ category because it does not provide descriptor-level information. The selection was stratified across 26 app categories~\cite{apple_app_genre}, taking the top 30 apps per category, where “top” is determined by sorting apps based on: (i) number of user ratings, (ii) average star rating, and (iii) popularity rank respectively. We argue that highly downloaded applications are less likely to contain incorrect information, as their large user base increases the likelihood of detecting and reporting violations, thereby enhancing the reliability of their content rating labels. For each app, we extract textual information such as short and long app descriptions, visual information such as the app icon and app screenshots, and developer-defined metadata such as content rating label and content rating descriptors. 
To evaluate our pipeline, we construct a iOS test set by selecting the next ten most popular apps per category for each content-rating class, yielding a dataset of 981 applications. 


%% file: sections/4_Methodology.tex
\section{Methodology}
\begin{figure*}[t]
    \centering
    \includegraphics[width=0.98\linewidth]{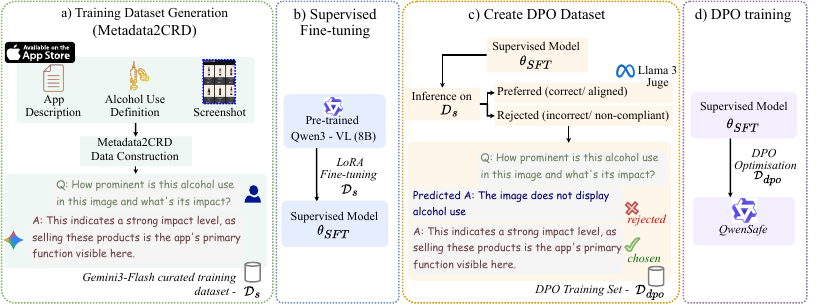}
    \caption{The overview of the QwenSafe pipeline. The pipeline involves four stage process. a) \textbf{metadata2CRD:} constructing QwenSafe training data from Apple app metadata, b) supervised fine tuning Qwen3-VL model, c) DPO dataset generation and d) DPO optimisation.}
    \label{fig:qwensape_pipeline}
    \vspace{-0.4cm}
\end{figure*}

In this section, we outline the overall methodology of QwenSafe, which comprises two primary components: (1) training data construction (Figure~\ref{fig:qwensape_pipeline}(a and c)) and (2) offline model adaptation for content-rating descriptor reasoning (Figure~\ref{fig:qwensape_pipeline}(b and d)). As detailed in Section~\ref{sec:training_data_generation}, we construct descriptor-aligned question–answer pairs from app metadata to provide supervision for CRD interpretation.
Section~\ref{sec:qwensafe_training} then describes the two-stage training pipeline, where the Qwen3-VL-8B model is adapted through LoRA-based supervised fine-tuning followed by mistake-driven Direct Preference Optimisation to align the model with descriptor-specific rationales.
\vspace{-0.2cm}

\subsection{Descriptor-Aligned Q\&A dataset Construction}
\label{sec:training_data_generation}
\noindent{\bf metadata2CRD:} As outlined in Section~\ref{chap3_datasets}, our unified descriptor taxonomy contains 30 fine-grained content descriptors.
Apple’s taxonomy provides only 12 coarse-grained descriptors, often aggregating multiple fine-grained concepts (\textbf{cf.}~Figure~\ref{fig:cd_taxonomy}(b)). For example, Apple provides only one aggregated descriptor for alcohol, tobacco, or drug use, despite the presence of six separate sub-descriptors in our taxonomy. To bridge this gap, we map Apple-provided labels onto our hierarchical taxonomy, enabling a consistent representation within a unified CRD space.

We further construct a descriptor-definition corpus by aggregating formal guidelines from ESRB, ACB, and Apple~\cite{esrb,acb,apple_new_cr_update}, associating each descriptor with a precise textual definition. For each app, we treat screenshots and developer-provided descriptions as multimodal inputs.

Building on this, we introduce metadata2CRD, a structured data generation procedure that produces descriptor-aligned question–answer pairs conditioned on visual content, textual metadata, and descriptor definitions. Specifically, given an app screenshot, its description, and a target descriptor with its definition, we prompt Gemini-2.5-Flash~\cite{comanici2025gemini} to generate reasoning traces that determine whether the descriptor is present and justify the decision using supporting visual and textual evidence. These generated rationales act as secondary CRD annotations, improving interpretability and serving as supervision signals for downstream training. Formally, metadata2CRD consists of four sequential reasoning steps:

\begin{itemize}
    \item Visual Understanding - The model identifies salient visual cues from the app screenshot that can be leveraged to characterise the content.
    \item Textual Understanding - The app description is analysed to build an understanding of the app’s intent and functionality, complementing the visual evidence.
    \item Descriptor-Definition Matching - Using the given descriptor definitions, the model evaluates whether the combined visual and textual evidence satisfies the criteria for the target descriptor.
    \item Q\&A Generation - The model summarises the above analyses into structured question–answer pairs that:(a) indicate the presence or absence of the descriptor in the image, and (b) justify this decision by citing relevant cues from both modalities.
\end{itemize}
An overview of the workflow can be found in Figure~\ref{fig:qwensape_pipeline}(a). From the secondary CRD tags generated by Gemini, we retain only those conversations whose predicted descriptors are consistent with the developer-defined labels. After filtering, our final training corpus comprises 79,500 high-quality Q\&A pairs with corresponding images.

\subsection{Offline Training of QwenSafe}
\label{sec:qwensafe_training}

\noindent {\bf Training:} Our training pipeline consists of two stages: (1) LoRA-based supervised fine-tuning (SFT) (\textbf{cf.}~Figure~\ref{fig:qwensape_pipeline}(b)) and (2) mistake-driven Direct Preference Optimization (DPO)(\textbf{cf.}~Figure~\ref{fig:qwensape_pipeline}(d)). 
We fine-tune the Qwen3-VL-8B-Instruct model using all filtered image–text Q\&A pairs generated through our metadata2CRD procedure. Because each training example is produced by jointly reasoning over screenshots, app descriptions, and descriptor definitions, the resulting corpus provides highly structured, descriptor-aligned supervisory signals. During SFT, the model is trained to minimize the standard next-token cross-entropy loss:

\begin{equation}
\label{eq:sft_loss}
\mathcal{L}_{\text{SFT}} 
= - \frac{1}{T}\sum_{t=1}^{T} \log p_{\theta} \left( y_t \mid x, y_{<t} \right),
\end{equation}

where T represents the number of output tokens, $y_{<t} = {y_1, ..., y_{t-1}}$ represents all the output tokens before position $t$ and $x$  represents the input sequence containing both visual tokens and text prompts. $p_{\theta}$ is the probability that the model assigns to the next token $y_t$ given all previous tokens $y_{<t}$, under the current parameters $\theta$. In this phase, the vision encoder and base language model remain fixed, and only the multimodal projection layer and LoRA adapters on selected transformer blocks are updated. This parameter-efficient strategy enables the VLM to learn CRD-specific reasoning while preserving the pre-trained visual and linguistic representations. 

Following the SFT phase, we employ DPO~\cite{rafailov2023direct} to further align the model with the CRD and enhance predictive accuracy. After completing SFT, we run the fine-tuned model over the entire SFT corpus and sample multiple responses per instance across three inference passes. These samples are compared against the reference answers produced by our metadata2CRD procedure, enabling us to collect both aligned (correct) and misaligned (incorrect) outputs. To identify misaligned responses, we employ Meta-Llama-3-8B-Instruct as a judge. For each item, the judge model is prompted with the original reference answer and the SFT-generated response and asked to assign an alignment score from 0 to 5 (0 as complete disagreement and 5 as perfect agreement). Responses with scores below 3 are treated as erroneous outputs and selected for DPO training. Each DPO pair thus consists of: (i) the ground-truth answer from the metadata2CRD process as the winning response, (ii) the misaligned SFT-generated answer as the losing response, and (iii) the associated visual-textual input(\textbf{cf.}~Figure~\ref{fig:qwensape_pipeline}(c)). The model is then optimized via the standard DPO objective:

\begin{equation}
\mathcal{L}_{\text{DPO}}(\pi_{\theta}; \pi_{\text{ref}})
= - \mathbb{E}_{(x, y_w, y_l) \sim \mathcal{D}}
\left[
    \log \sigma \left(\hat{r}_{\theta}(x, y_w)-\hat{r}_{\theta}(x, y_l)\\
    \right)
\right]
\end{equation}
\begin{equation*}
\text{where }
\hat{r}_{\theta}(x, y) = \beta \log \frac{\pi_{\theta}(y \mid x)}{\pi_{\text{ref}}(y \mid x)}
\end{equation*}


Here, $\pi_{\theta}$ denotes the model being optimized, $\pi_{\text{ref}}$ is the SFT model, and $y_w$ and $y_l$
represent the preferred and dis-preferred responses, respectively. This stage prioritises the model to generate more descriptor-consistent rationales. 

%% file: sections/5_Results.tex
\section{Experimental Setup}
\noindent {\bf Dataset:} Using the procedure described in Section~\ref{sec:training_data_generation}, we construct a corpus of 79,500 descriptor-aligned question–answer pairs derived from 2,340 iOS applications. This dataset serves as the training dataset for the SFT of the QwenSafe model. Following SFT, we further obtain 7,356 mistake-focused image–text preference pairs, which are used to train the model through DPO. We used 981 iOS apps as the test dataset to evaluate the QwenSafe model.

\noindent {\bf Evaluation Metrics:} 
The goal of QwenSafe is to reliably detect the presence of specific content rating descriptors in mobile app metadata. To assess model performance, we formulate the task as 1) binary classification for each CRD, where the model’s generated textual output is mapped to a positive or negative label and 2) multi-class classification, which predicts the CRD impact as either strong or mild (see Section~\ref{sec:ios_results} for details). For each descriptor, we report precision and recall for both the positive and negative classes. Among these metrics, we prioritise \textit{positive recall} and \textit{negative precision}, as they capture the two most critical behaviours for safety rating: (i) high positive recall ensures that instances containing objectionable or rating-relevant content are correctly identified, while (ii) high negative precision reduces false alarms by ensuring that benign content is classified as harmful. This aligns the evaluation with the practical requirements of content rating systems, which must avoid both missed violations and unnecessary over-flagging.  

\noindent {\bf Baselines:} To evaluate the performance of QwenSafe, we compare against several strong vision–language models (VLMs). Specifically, we evaluate: (i) Qwen3-VL-8B~\cite{Qwen3-VL}, the underlying backbone of QwenSafe; (ii) LLaVA-v1.6-Mistral-7B~\cite{liu2023improved}, a similarly sized open-source VLM capable of joint image–text reasoning; and (iii) Gemini-2.5-Flash~\cite{comanici2025gemini}, a state-of-the-art multimodal model from Google DeepMind. Gemini-2.5-Flash is included both as a high-capacity reasoning baseline and because it is used in our data construction pipeline for generating SFT training corpora. For each model, we provide the app screenshot and its corresponding textual description as input, and prompt the model to determine the presence or absence of each content rating descriptor. This setup ensures a consistent and comparable evaluation across all baselines.
\vspace{-0.3cm}
\subsection{Evaluation Results}
\begin{figure*}[t]
    \centering
    \includegraphics[width=0.9\linewidth]{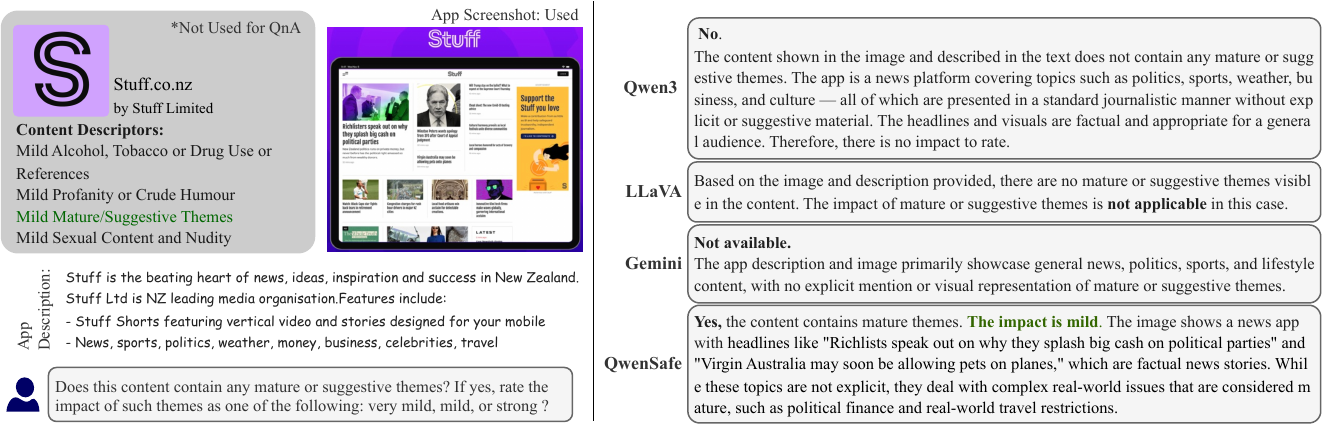}
    \caption{Example illustrating model behaviour on the Mature/Suggestive Themes descriptor. QwenSafe recognises the subtle cues present in both the screenshot and description and accurately labels the impact as mild, demonstrating improved sensitivity to low-intensity content. }
    \label{fig:qwensafe_example}
    \vspace{-0.3cm}
\end{figure*}

\label{sec:ios_results}
\begin{table*}[t]
\caption{Binary classification performance comparison. Positive class recall ($R^+$) and negative precision ($P^-$) across content rating descriptors for LLaVA-v1.6-mistral-7b-hf~\cite{liu2023improved}, Qwen3-VL-8B~\cite{Qwen3-VL}, Gemini-2.5-Flash~\cite{comanici2025gemini}, and QwenSafe(ours). Best performance is denoted in bold.(values in percentage).}
\centering
\footnotesize
\begin{tabular}{lcccccccccc}
\hline
\textbf{Content Rating Descriptor}&\multicolumn{2}{c}{\textbf{LLaVA}} &
\multicolumn{2}{c}{\textbf{Qwen3}} &
\multicolumn{2}{c}{\textbf{Gemini}} &
\multicolumn{2}{c}{\textbf{QwenSafe}} \\
 & \textbf{$R^+$} & \textbf{$P^-$} & \textbf{$R^+$} & \textbf{$P^-$} & \textbf{$R^+$} & \textbf{$P^-$} & \textbf{$R^+$} & \textbf{$P^-$}\\
\hline
Alcohol, Tobacco or Drug &  &  &  &  &  &  &  &  \\
Use or References & 16.87 & 85.52 & 13.86 & 85.07 & 43.37 & 85.22 & \textbf{54.22} & \textbf{91.15}  \\
Horror or Fear Themes                     & 45.22 & \textbf{93.22} & 26.09 & 91.06 & \textbf{48.70} & 90.95 & 40.00 & 92.38  \\
Mature or Suggestive Themes               & 16.25 & 74.65 & 24.38 & 76.54 & 52.30 & 75.89 & \textbf{55.48} & \textbf{79.78} \\
Medical / Treatment Information           & 45.05 & 93.45 & 36.94 & 92.55 & 55.86 & 92.41 & \textbf{57.66} & \textbf{94.33}  \\
Profanity or Crude Humour                 & 22.87 & 84.54 & 10.11 & 82.43 & 41.49 & 82.54 & \textbf{46.28} & \textbf{88.55} \\
Cartoon or Fantasy Violence               & 33.96 & 88.67 & 28.30 & 87.82 & \textbf{55.97} & 88.69 & 52.20 & \textbf{91.10}  \\
Realistic Violence                        & 34.25 & 94.98 & 26.03 & 94.39 & 52.05 & 94.78 & \textbf{60.27} & \textbf{96.64}  \\
Sexual Content or Nudity                  & 33.33 & 90.16 & 27.54 & 89.40 & 42.75 & 87.83 & \textbf{51.45} & \textbf{92.29}  \\
Simulated Gambling                        & \textbf{40.54} & \textbf{97.72} & 16.22 & 96.82 & 32.43 & 96.41 & 35.14 & 97.44\\
Real Gambling                             & 15.79 & 95.06 & 12.28 & 94.87 & 15.79 & 93.13 & \textbf{24.56} & \textbf{95.46}  \\
Unrestricted Web Access                   & 74.29 & 97.01 & 16.19 & 90.87 & \textbf{87.62} & \textbf{97.89} & 60.95 & 93.84  \\
Contests                                  & 71.79 & \textbf{98.85} & 51.28 & 98.02 & 71.79 & 98.38 & \textbf{74.36} & 98.62 \\
\hline
\textbf{Average} & 37.52 & 91.15 & 24.10 & 89.99 & 50.01 & 90.34 & \textbf{51.05} & \textbf{92.63} \\
\hline
\end{tabular}
\label{tab:ios_results_percentage}
\end{table*}

\begin{table}[t]
\caption{Multi-class classification: Average precision and recall across nine different content rating descriptors. We report mild and strong average precision and recall for Qwen3-VL-8B~\cite{Qwen3-VL}, Gemini-2.5-Flash~\cite{comanici2025gemini}, and QwenSafe. The best recall for both classes is recorded in bold (values in percentage).}
\centering
\footnotesize
\begin{tabular}{lcccc}
\hline
\textbf{Model} & \textbf{$P_{mild}$} & \textbf{$R_{mild}$} & \textbf{$P_{strong}$} & \textbf{$R_{strong}$} \\
\hline
Qwen3    & 73.98 & 4.58  & 40.74 & 45.91 \\
Gemini   & 83.94 & 20.91 & 30.04 & 49.05 \\
QwenSafe & 49.50 & \textbf{34.03} & 27.31 & \textbf{55.25} \\
\hline
\end{tabular}
\label{tab:multiclass_avg}
\vspace{-0.5cm}
\end{table}




\noindent \textbf{Binary classification.}
We evaluate the effectiveness of QwenSafe in detecting 12 different safety-critical CRDs from Apple, framed as a binary verification task over app metadata (images and descriptions). A descriptor is considered present if any image–text pair indicates its occurrence, reflecting a conservative setting for safety auditing. Figure~\ref{fig:qwensafe_example} illustrates that QwenSafe correctly identifies mature suggestive themes with supporting rationale, whereas general-purpose VLMs fail to detect their presence.
As shown in Table~\ref{tab:ios_results_percentage}, QwenSafe consistently outperforms baseline VLMs. Compared to Qwen3-VL-8B, QwenSafe improves positive-class recall ($R^+$) by 111.8\%, indicating substantially better coverage of descriptor-relevant content, while also increasing negative-class precision ($P^-$) by 2.9\%, reducing false positives. Relative to LLaVA-v1.6-7B, QwenSafe achieves a 36.1\% gain in $R^+$, highlighting its effectiveness in detecting safety-critical signals. 
Both LLaVA and Qwen3 exhibit lower recall and comparatively high negative precision, suggesting a tendency to under-predict descriptor presence. This reflects the limitations of general-purpose VLMs for CRD-specific reasoning. Gemini-2.5-Flash performs competitively (50.01\% $R^+$, 90.3\% $P^-$), yet QwenSafe surpasses it by at least 2.1\% on both metrics, demonstrating stronger alignment with descriptor-specific cues.

\noindent \textbf{Multi-class classification.}
Table~\ref{tab:multiclass_avg} reports average precision and recall for mild and strong severity levels across nine descriptors (excluding gambling, unrestricted web access, and contests, as Apple does not provide severity annotations). Full per-category results are provided in Appendix Table~\ref{tab:appendixmulticlass}.

QwenSafe achieves the best overall performance across both severity levels. In particular, it improves mild-class recall by 62.75\% relative to Gemini and attains the highest strong-class recall (0.5525), exceeding Gemini by 12.64\%. In contrast, Qwen3-VL exhibits the lowest recall for both mild class ($R_{\text{mild}}=4.58\%$) and strong class($R_{\text{strong}}=45.91\%$), comparatively the recall of QwenSafe is at least $20.3\%$ higher. This trend is consistent with its binary performance, where limited recall suggests weak sensitivity to fine-grained CRD cues. However, both Gemini and Qwen3 show limited ability to distinguish between severity levels, suggesting that general-purpose VLMs capture coarse content signals but lack alignment for graded descriptor reasoning.

Whereas QwenSafe provides consistently higher recall across both mild and strong classes, indicating improved sensitivity to descriptor presence and severity. Although mild cases remain challenging—particularly when metadata lacks explicit cues—QwenSafe maintains robust performance. 
\textit{These results demonstrate that QwenSafe enables fine-grained, reliable identification of harmful app content through metadata2CRD-driven training, supporting automated enforcement of platform safety policies in mobile ecosystems.}
\vspace{-0.4cm}

\subsection{Non-Disclosed Descriptors.}
\begin{figure*}[t]
    \centering
    \includegraphics[width=0.95\linewidth]{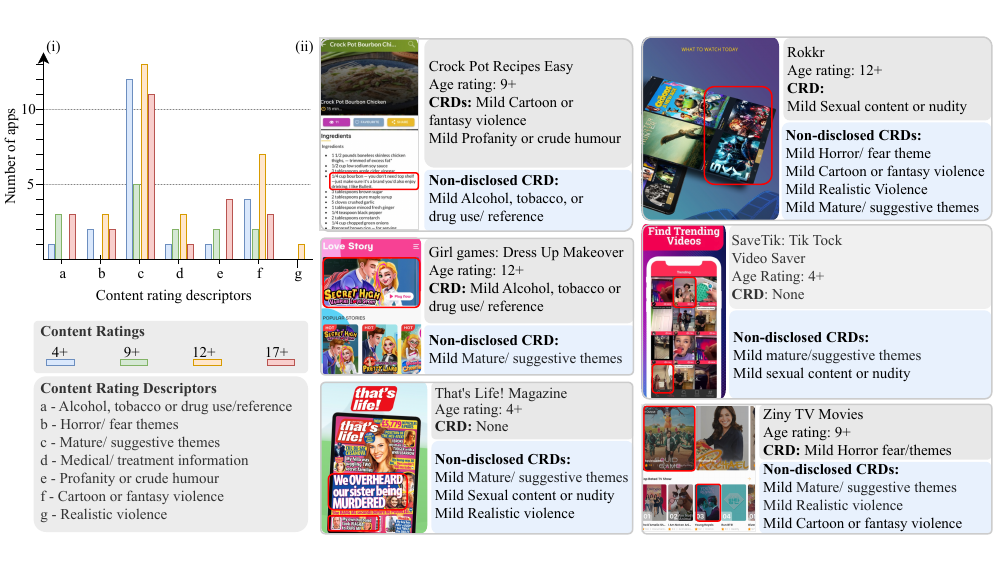}
    \caption{Analysis of non-disclosed CRDs identified by QwenSafe. (i) Distribution of applications containing non-disclosed CRDs across Apple age rating categories (4+, 9+, 12+, and 17+) and descriptor types. (ii) Representative examples of applications where QwenSafe detects CRDs that are not declared in the app metadata.}
    \label{fig:non_discloased_examples}
    \vspace{-0.5cm}
\end{figure*}
Apple’s age-rating system requires developers to disclose the presence of specific content rating descriptors (CRDs). However, some descriptors may appear in the app’s visual content but remain unreported in the app metadata. We refer to these as non-disclosed descriptors. To quantify and analyse such cases, we evaluate all apps in our test set using QwenSafe and assess the presence of all 12 Apple-defined CRDs, regardless of what developers reported. Figure~\ref{fig:non_discloased_examples} (i) summarises the number of apps containing non-disclosed CRDs for each descriptor, while Figure~\ref{fig:non_discloased_examples} (ii) presents representative examples identified by QwenSafe. According to Apple’s guidelines, apps rated 4+ are expected to contain no objectionable material. However, our analysis reveals 21 apps within the 4+ category that exhibit at least one non-disclosed CRD. Notably, Apple’s age rating definitions specify that apps containing mild cartoon or fantasy violence, mild profanity or crude humour, or mild mature, suggestive, or horror-themed content must be rated 9+ or higher. As illustrated in Fig.~\ref{fig:non_discloased_examples}(ii), we identify multiple 4+ applications that display visible mature or suggestive themes, rendering them inappropriate for the intended age group.
We further observe similar inconsistencies in higher age categories. Specifically, within the 9+ category, we identify 14 applications with non-disclosed CRDs. These include applications with alcohol references as well as a media streaming application for which QwenSafe detects additional descriptors, including realistic violence, that are not reflected in the declared metadata. While Apple permits mild or infrequent mature or suggestive themes for the 9+ category, we identify five applications in which such content is present but not disclosed. Comparable patterns of missing descriptors are also observed in applications rated 12+ and 17+.

Apple states that applications undergo manual review prior to publication on the App Store. Nevertheless, our evaluation indicates that 9.37\% of applications in the test set contain at least one non-disclosed CRD. As discussed in Sec.~\ref{sec:ios_results} and illustrated in Fig.~\ref{fig:qwensafe_example}, even state-of-the-art vision–language models such as Gemini and Qwen3 exhibit limited effectiveness in reliably identifying CRDs and their severity levels. This gap introduces potential safety risks, particularly for younger users. \emph{In contrast, QwenSafe demonstrates the ability to systematically detect undisclosed descriptors and align predictions with Apple’s content rating definitions, providing a more reliable and policy-aligned solution for this downstream safety-critical task.}
\vspace{-0.4cm}

%% file: sections/6_Conclusion.tex
\section{Conclusion}
We introduce QwenSafe, a vision–language model for automatic identification of safety-critical content signals in mobile applications, supported by metadata2CRD, a scalable data construction pipeline that generates descriptor-aligned supervision from multimodal app metadata without relying on static and runtime behaviours. By combining supervised fine-tuning with DPO-based preference alignment, QwenSafe consistently outperforms strong vision–language baselines across 12 Apple-defined descriptors, with substantial gains in positive-class recall, indicating improved detection of potentially harmful or misreported content by app developers. These results demonstrate the effectiveness of descriptor-aware multimodal alignment for automated content classification and highlight the potential of vision–language models to support automated, scalable enforcement of safety and compliance in mobile ecosystems. 

%% file: sections/7_appendix.tex
\section{Appendix}
\label{sec:appendix}

Due to space constraints, we provide the complete multi-class classification results across all descriptors in Table~\ref{tab:appendixmulticlass}. This table reports mild and strong precision and recall for all methods, complementing the summarised results in Section~\ref{sec:ios_results}.
\begin{sidewaystable}
\centering
\scriptsize
\caption{Multi-class classification performance across content rating descriptors. We report mild and strong precision and recall for Qwen3-VL-8B~\cite{Qwen3-VL}, Gemini-2.5-Flash~\cite{comanici2025gemini}, and QwenSafe. We exclude gambling, unrestricted web access, and contests because Apple does not provide impact levels for these descriptors. The best recall for both classes is recorded in bold.}
\begin{tabular}{lcccc|cccc|cccc}
\hline
\textbf{Content Rating Descriptor} &
\multicolumn{4}{c}{\textbf{Qwen3}} &
\multicolumn{4}{c}{\textbf{Gemini}} &
\multicolumn{4}{c}{\textbf{QwenSafe}} \\
 & \textbf{$P_{mild}$} & \textbf{$R_{mild}$} & \textbf{$P_{strong}$} & \textbf{$R_{strong}$} &
   \textbf{$P_{mild}$} & \textbf{$R_{mild}$} & \textbf{$P_{strong}$} & \textbf{$R_{strong}$} &
   \textbf{$P_{mild}$} & \textbf{$R_{mild}$} & \textbf{$P_{strong}$} & \textbf{$R_{strong}$} \\
\hline
Alcohol, Tobacco or Drug  &&&&&&&&&&&\\
Use/ References & 71.43 & 6.76 & 66.67 & 33.33 & 92.59 & 33.78 & 50.00 & \bf50.00 & 69.52 & \textbf{49.32} & 52.94 & \bf50.00 \\
Horror or Fear Themes                     & 75.00 & 5.77 & 22.73 & 45.45 & 76.19 & 15.38 & 11.43 & 36.36 & 50.00 & \bf23.08 & 28.57 & \bf72.73 \\
Mature or Suggestive Themes               & 88.89 & 3.48 & 55.00 & \bf62.26 & 80.77 & 27.39 & 35.71 & 47.17 & 31.95 & \bf36.96 & 31.52 & 54.72 \\
Medical / Treatment Information           & 100.00 & 3.37 & 44.74 & \bf77.27 & 80.00 & 17.98 & 38.10 & 72.73 & 26.55 & \bf33.71 & 35.90 & 63.64 \\
Profanity or Crude Humour                 & 90.00 & 5.45 & 33.33 & 13.04 & 82.93 & 20.61 & 16.22 & \bf26.09 & 81.11 & \bf44.24 & 22.22 & 8.70 \\
Cartoon or Fantasy Violence               & 90.48 & 13.29 & 25.00 & 37.50 & 85.11 & 27.97 & 16.67 & 43.75 & 54.22 & \bf31.47 & 20.45 & \bf56.25 \\
Realistic Violence                        & 50.00 & 1.52 & 5.88 & 14.29 & 90.00 & 13.64 & 10.71 & \bf42.86 & 35.59 & \bf31.82 & 5.17 & \bf42.86 \\
Sexual Content or Nudity                  & 100.00 & 1.63 & 33.33 & \bf80.00 & 92.86 & 21.14 & 29.03 & 60.00 & 63.24 & \bf34.96 & 25.00 & 73.33 \\
Simulated Gambling                        & 0.00 & 0.00 & 80.00 & 50.00 & 75.00 & 10.34 & 62.50 & 62.50 & 33.33 & \bf20.69 & 24.00 & \bf75.00 \\
\hline
\textbf{Average} & 73.98 & 4.58 & 40.74 & 45.91 & 83.94 & 20.91 & 30.04 & 49.05 & 49.50 & \textbf{34.03} & 27.31 & \textbf{55.25} \\
\hline
\end{tabular}
\label{tab:appendixmulticlass}

\end{sidewaystable}